# View-adaptive Renderer for View-consistent 2D-to-3D Generation

U-Chae Jun[1†], Jaeeun Ko[1†], Jiwoo Kang[1,2*]

[1]Department of IT Engineering, Sookmyung Women's University, Seoul, 04310, South Korea.
[2]Division of Artificial Intelligence Engineering, Sookmyung Women's University, Seoul, 04310, South Korea.

*Corresponding author(s). E-mail(s): jwkang@sookmyung.ac.kr;
Contributing authors: wjsdbco@sookmyung.ac.kr;
rhwodms1223@sookmyung.ac.kr;
†These authors contributed equally to this work.

**Abstract**

Reconstructing 3D shapes from a single image remains a fundamental yet challenging problem in computer vision. Traditional monocular 3D generation pipelines typically synthesize multiple views from a single input image before applying Neural Radiance Field (NeRF)-based reconstruction. However, inherent projective ambiguities often produce visual discontinuities across generated viewpoints, leading to inaccuracies in reconstructed 3D models. Current solutions either incur significant additional computational burdens or fail to adequately resolve practical inconsistencies between synthesized views. To address these limitations, we propose a novel *viewpoint-adaptive neural rendering* framework that enables robust 3D reconstruction even when given partially inconsistent multi-view inputs. Our approach introduces view-adaptive neural renderers that independently correct viewpoint-dependent errors while simultaneously sharing a global feature backbone to preserve structural coherence. Furthermore, we propose a self-attention fusion module that adaptively integrates multi-view information, ensuring geometric consistency without relying heavily on indirect regularizations or computationally intensive methods. Through extensive experiments, we demonstrate that our method consistently improves 3D reconstruction fidelity. Importantly, our approach achieves near state-of-the-art performance without diffusion-based SDS supervision, relying primarily on photometric rendering loss with lightweight attention regularizers. This balance between accuracy and efficiency makes the proposed framework highly practical for real-world applications.

# 1 Introduction

Humans possess an innate ability to infer the 3D structure of objects from a single 2D image by leveraging geometric priors such as symmetry, occlusion, and perspective cues. Inspired by this capability, recent deep learning-based approaches have made remarkable progress in single-image 3D reconstruction by incorporating data-driven priors and generative models [1–3]. Despite these advancements, achieving accurate 3D reconstructions from a single image remains a challenging problem due to the fundamental ambiguity in projecting a 2D image into a 3D space.

A single 2D photograph inherently discards critical depth and surface information, making the reconstruction of complex 3D geometry an ill-posed problem. Moreover, recent methods typically rely on generating multiple views from a single input image and then apply NeRF-based rendering [4–6]. However, *monocular multi-view image generation* is itself prone to inconsistencies across different camera angles due to perspective ambiguity, limited context, or imperfect generative priors. These inconsistencies directly undermine the quality of downstream 3D reconstructions, resulting in blurred surfaces or distorted geometry. Hence, a key motivation of our work is to develop a method that *explicitly accounts for viewpoint-dependent discrepancies* and seamlessly fuses local corrections without sacrificing global coherence.

To address these challenges, a common paradigm for single-image-to-3D reconstruction involves a two-stage process: (1) generating multi-view images from the given single image and (2) reconstructing the 3D structure using NeRF-based methods. The success of this pipeline heavily depends on the quality and consistency of the generated multi-view images. However, due to projective ambiguity, monocular image-based multi-view generation methods often produce images with significant discrepancies across different viewpoints. These inconsistencies lead to artifacts and distortions in the final 3D reconstruction, limiting the reliability of existing approaches.

Several methods have been proposed to mitigate view inconsistency in multi-view images. One line of research attempts to enhance multi-view synthesis consistency by leveraging synchronized generative models [2] or explicit geometric constraints [7]. While these approaches improve view alignment, they remain fundamentally limited by the capacity of the image generation model, and perfect consistency remains unattainable. Another line of research focuses on refining 3D reconstruction by integrating self-supervised regularization techniques such as Score Distillation Sampling (SDS) [7] and Contrastive Language-Image Pretraining (CLIP) [8]. These methods provide implicit constraints to guide the reconstruction process, but they introduce high computational overhead and often fail to resolve local inconsistencies in generated views.

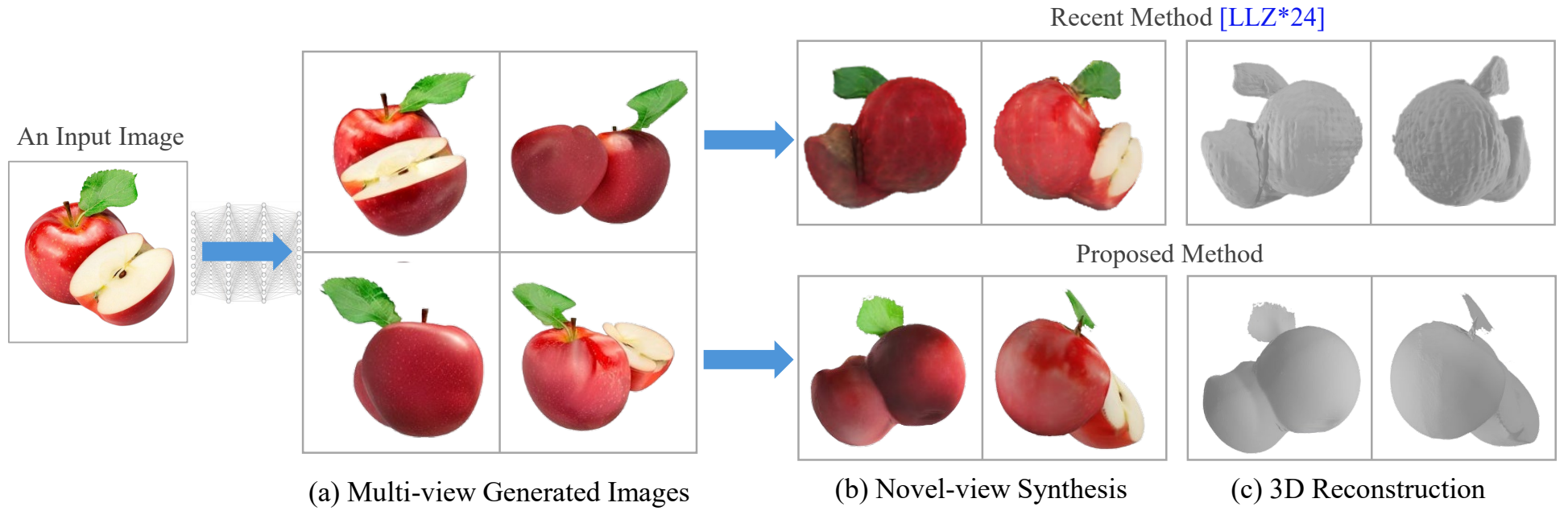

**Fig. 1** An overview of our single image to 3D reconstruction framework. (a) Multi-view images are generated from a single input using state-of-the-art multi-view generation methods [2]. (b) A variant of *NeRF* is applied for novel-view synthesis, and (c) these synthesized views are employed for 3D reconstruction. Due to projective ambiguity, the generated multi-view images often exhibit visual discontinuities that compromise reconstruction quality. To address this challenge, we introduce view-adaptive neural renderers that effectively correct viewpoint-dependent inconsistencies, thereby achieving accurate and coherent 3D reconstructions.

To address these limitations, we propose a novel *view-adaptive neural rendering framework* that improves 3D reconstruction accuracy by explicitly learning viewpoint-dependent representations. Instead of relying solely on the consistency of multi-view images, our method incorporates *view-adaptive neural renderers* that specialize in processing individual viewpoints while maintaining shared global representations. This approach allows the model to adaptively correct inconsistencies across different viewpoints and produce geometrically coherent 3D reconstructions. Additionally, we introduce a *self-attention fusion module* that integrates learned features across multiple views, ensuring robustness to partial view inconsistencies.

Our main contributions can be summarized as follows:

- We introduce a *view-adaptive neural renderer* that enables more accurate 3D reconstruction from partially inconsistent multi-view images by assigning specialized network layers to each viewpoint.
- We propose a *self-attention fusion mechanism* that adaptively integrates viewpoint-adaptive features while preserving global structural coherence.
- We conduct extensive experiments showing that the proposed method improves both novel-view synthesis and 3D reconstruction quality over NeRF-based baselines under shared-backend evaluation settings.

Experimental results show that our approach not only enhances reconstruction fidelity but also reduces computational overhead compared to alternative strategies that rely on indirect supervision. Our method thus offers a robust and efficient solution for single-image 3D generation, with potential applications in computer graphics, virtual reality, and content creation.

## 2 Related Work

In this section, we review prior research related to single-image-to-3D reconstruction, multi-view image generation, and neural rendering. Our discussion highlights the key challenges in multi-view consistency and how our approach differs from existing methods.

### 2.1 Single-Image 3D Reconstruction

Recovering 3D structures from a single image has been a longstanding challenge in computer vision. Early methods relied on geometric constraints, such as structure-from-motion (SfM) [9] and shape-from-shading [10]. These approaches required either multiple views or strong assumptions about lighting and material properties, limiting their applicability to real-world scenes.

Deep learning-based methods have significantly advanced single-image 3D reconstruction by leveraging large datasets and powerful neural networks. Voxel-based approaches, such as 3D-R2N2 [11] and Pix2Vox [12], predict discrete voxel grids from input images using convolutional networks. While effective, their resolution is limited by the cubic memory complexity of voxel representations. Mesh-based methods, such as AtlasNet [13] and Pixel2Mesh [14], directly predict 3D mesh structures, allowing for high-resolution reconstructions but struggling with complex topologies. More recent implicit function-based approaches, such as DeepSDF [15] and Occupancy Networks [16], represent 3D surfaces as continuous functions, enabling smooth reconstructions. These methods have been extended into neural radiance field (NeRF) frameworks [4] to model view-dependent radiance.

Despite these advancements, single-image methods often produce reconstructions that lack fine-grained multi-view consistency. This limitation motivates our work, which explicitly models viewpoint-dependent variations to correct inconsistencies in multi-view images.

### 2.2 Multi-View Image Generation

Generating multiple views from a single image is a crucial step in monocular 3D reconstruction. Traditional methods estimate depth from a single image [17] and use geometric warping [18] to synthesize new views. However, these methods struggle with occlusions and texture ambiguities, often resulting in unrealistic or distorted images.

Recent generative models have improved multi-view synthesis by incorporating strong data-driven priors. Diffusion-based models, such as Score Distillation Sampling (SDS) in DreamFusion [7], refine 3D representations by iteratively optimizing a radiance field. SyncDreamer [2] enhances multi-view consistency through synchronized diffusion processes, while Zero-1-to-3 [19] and Wonder3D [20] further improve novel view generation by enforcing explicit geometric constraints.

Another line of research leverages pre-trained 2D priors to lift single images into a 3D-consistent space. Magic3D [21] refines generated multi-view images through high-resolution supervision, while RealFusion [3] utilizes a 2D diffusion model to progressively refine multi-view images. Despite these improvements, achieving perfect multi-view consistency remains an open problem. Our approach complements these advancements by introducing a neural rendering framework that explicitly learns viewpoint-dependent representations, ensuring geometric coherence even when generated views contain inconsistencies.

Our work differs from prior studies such as RealFusion [3] and Zero-1-to-3 [19] in that we explicitly address view inconsistencies through view-adaptive adaptation rather than relying solely on global loss functions. This targeted approach enables more robust 3D reconstructions even when the input views exhibit significant disparities.

### 2.3 Neural Rendering and View Consistency

Neural rendering has revolutionized 3D reconstruction by enabling photorealistic novel view synthesis. NeRF-based methods [4] have demonstrated impressive results in synthesizing novel views from multi-view images by modeling volumetric radiance fields. However, the effectiveness of NeRF heavily depends on the consistency of input images.

Several techniques have been proposed to address multi-view inconsistency. Multi-scale feature learning, as seen in Mip-NeRF [22], reduces aliasing artifacts by encoding hierarchical representations. Camera-aware optimization methods, such as NeRF++ [23] and FreeNeRF [24], refine camera parameters alongside scene reconstruction to mitigate errors from misaligned views. Regularization techniques, such as MVDream [25], apply 3D self-attention to enforce geometric consistency, while FreeNeRF [24] introduces frequency-based constraints to improve novel view synthesis with limited observations.

Despite these improvements, existing NeRF-based models assume a fixed scene representation across all viewpoints, making them susceptible to artifacts when input images contain inconsistencies. Our method introduces a view-adaptive neural rendering framework that explicitly models viewpoint-dependent variations, enabling adaptive corrections and robust 3D reconstruction.

### 2.4 Self-Attention in Neural Rendering

Self-attention mechanisms have been widely used in deep learning to capture long-range dependencies [26]. In neural rendering, self-attention has been explored in various forms to enhance 3D synthesis. EG3D [27] integrates self-attention with GAN-based 3D synthesis, while DreamAvatar [28] employs cross-attention mechanisms to refine neural avatars. More recently, Dual Pose-invariant Embeddings [29] propose learning category- and object-specific discriminative representations for multi-view object recognition, improving the alignment of multi-view features.

Inspired by these works, we introduce a self-attention fusion module that adaptively integrates viewpoint-adaptive features. Unlike existing approaches, which primarily focus on improving neural rendering quality through more expressive feature representations, our self-attention mechanism is explicitly designed to correct inconsistencies in generated multi-view images while preserving global geometric coherence.

# 3 View-adaptive Neural Rendering

In this section, we introduce our view-adaptive neural rendering framework designed to overcome partial view inconsistencies in multi-view images. We begin with an **Overview** of our approach, outlining the core challenges and motivations behind our design. Next, we detail the **Network Architecture** by first discussing the limitations of standard multi-view encoding in NeRF and then presenting our novel view-adaptive neural representation that decouples local viewpoint errors from global scene features. We follow this with a comprehensive description of our **Self-Attention Fusion** module, which integrates view-adaptive features through attention-based aggregation and regularization techniques to ensure coherent 3D reconstructions. Afterward, we outline our multi-phase **Training Strategy** that balances global pre-training with view-adaptive fine-tuning. Finally, we provide a **Theoretical Analysis** that justifies how our framework mitigates view inconsistencies and improves reconstruction quality.

## 3.1 Overview

The fundamental motivation behind our method is to tackle the issue of multi-view inconsistency, a common obstacle in single-image 3D reconstruction pipelines. Existing *NeRF*-based methods [4–6, 22] often assume multi-view images are consistent, but real or generated images from monocular inputs can exhibit significant discrepancies across viewpoints. An overview of the proposed architecture is presented in Fig. 2.

To address this, we propose a framework in which:

1. A **shared feature extractor** captures global geometric and appearance priors.
2. A set of **view-adaptive neural renderers** is introduced to learn viewpoint-dependent variations, effectively "correcting" inconsistency for each view.
3. A **self-attention fusion** module adaptively aggregates multi-view information into a coherent global representation.

This design decouples local viewpoint errors from the global scene representation, ensuring that each view's unique discrepancies are addressed individually while preserving overall structural coherence. Furthermore, our framework significantly reduces training overhead by mitigating the need for heavy indirect regularization.

## 3.2 Network Architecture

The proposed *view-adaptive neural rendering* framework builds upon the standard NeRF formulation [4], but introduces key modifications to handle view inconsistency in multi-view images. This section first describes the limitations of traditional multi-view encoding in NeRF and then presents our view-adaptive representation learning strategy.

### 3.2.1 Standard Multi-View Encoding in NeRF

In standard NeRF-based models, a single MLP is trained to encode the radiance field for all viewpoints. Given a 3D point $\mathbf{x} = (x, y, z)$ and a viewing direction $\mathbf{\Theta} = (\theta, \phi)$,

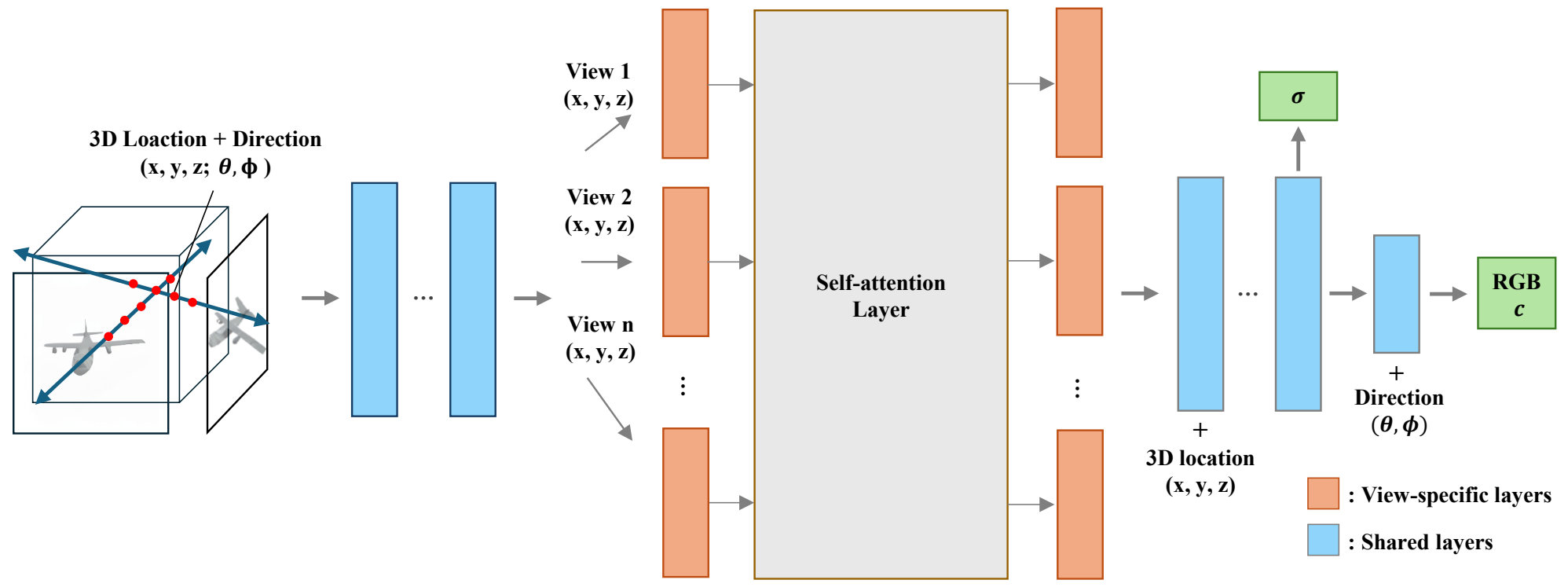

**Fig. 2** Network architecture of the proposed view-adaptive neural renderer. The self-attention module aggregates multi-view features to extract robust view-adaptive representations while preserving global coherence.

the model predicts the color $\mathbf{c} = (r, g, b)$ and volume density $\sigma$ as:

$$(\mathbf{c}, \sigma) = \mathcal{F}_{\text{NeRF}}(\mathbf{x}, \mathbf{\Theta}). \tag{1}$$

To enable high-frequency detail reconstruction, positional encoding $\gamma(\cdot)$ is applied to both the spatial coordinates and viewing directions [4]:

$$\mathbf{f} = \mathcal{F}_{\text{shared}}(\gamma(\mathbf{x}), \gamma(\mathbf{\Theta})). \tag{2}$$

Despite its effectiveness in view synthesis, this approach assumes that all viewpoints share a globally consistent representation. When multi-view images contain inconsistencies due to inaccuracies in monocular image generation or real-world artifacts, the learned representation becomes biased, leading to visual distortions and artifacts in reconstructed 3D geometry [2, 7]. The standard NeRF pipeline lacks a mechanism to model view-dependent variations separately, which limits its ability to correct inconsistencies in input views.

This limitation becomes especially significant when even minor viewpoint-adaptive discrepancies accumulate, resulting in amplified reconstruction errors. To mitigate this issue, our approach incorporates dedicated view-adaptive MLP branches that explicitly model and correct for these variations, thereby enhancing overall 3D reconstruction fidelity.

### 3.2.2 View-Adaptive Neural Representation

To address these limitations, we introduce *view-adaptive neural representations* that explicitly model viewpoint-dependent variations while maintaining a shared global structure. Instead of encoding all viewpoints with a single MLP, we decompose the network into a **shared feature extractor** and a set of **view-adaptive neural renderers**.

The shared feature extractor follows the standard NeRF encoding scheme:

$$\mathbf{f}_{\text{shared}} = \mathcal{F}_{\text{shared}}(\gamma(\mathbf{x}), \gamma(\boldsymbol{\Theta})). \tag{3}$$

Each viewpoint $i$ is then processed by a dedicated MLP:

$$\mathbf{f}_i = \mathcal{F}_{\text{view},i}(\mathbf{f}_{\text{shared}}), \quad i \in \{1, ..., n\}. \tag{4}$$

By allowing each viewpoint to have independent feature transformations, the model can adaptively correct local inconsistencies without corrupting the shared representation. In this way, this decoupling enables each view-adaptive branch to focus solely on rectifying the unique distortions of its corresponding view, thereby reducing error propagation between views. The view-adaptive features are then fused using a self-attention mechanism (detailed in Sec. 3.3) to ensure global coherence.

***Role of View-Adaptive MLPs:***
The dedicated MLP for each viewpoint is designed to learn corrections specific to the inconsistencies observed in that view. In practice, each branch can tailor its parameters to handle distinct noise patterns and misalignments inherent in the generated multi-view images. By processing the shared feature $\mathbf{f}_{\text{shared}}$ independently, each view-adaptive branch can adjust local features without corrupting the global representation, thereby mitigating artifacts due to view inconsistencies.

## 3.3 Self-Attention Fusion

To ensure that viewpoint-adaptive representations remain geometrically coherent, we introduce a *self-attention fusion module*. This module aggregates information across multiple viewpoints using a multi-head self-attention mechanism, allowing the model to selectively integrate reliable viewpoint information while suppressing inconsistencies. This fusion step is critical for compensating for noisy or misaligned view-adaptive features by leveraging complementary information from other views.

### 3.3.1 Attention Mechanism for View Aggregation

Given a set of view-adaptive features $\{\mathbf{f}_1, \mathbf{f}_2, ..., \mathbf{f}_n\}$ from the view-adaptive neural renderers, our goal is to compute a fused representation $\mathbf{f}_{\text{fused}}$ that retains global coherence while adapting to inconsistencies in individual viewpoints.

The self-attention mechanism is based on the standard transformer attention formulation [26]. Each view-adaptive feature $\mathbf{f}_i$ is first projected into query, key, and value representations:

$$\mathbf{q}_i = W_q\mathbf{f}_i, \quad \mathbf{k}_i = W_k\mathbf{f}_i, \quad \mathbf{v}_i = W_v\mathbf{f}_i. \tag{5}$$

Here, the projection matrices $W_q$, $W_k$, and $W_v$ are learnable parameters that adapt the attention mechanism to the characteristics of the view-adaptive features.

The attention weight $\alpha_{ij}$ between viewpoints $i$ and $j$ is computed using the scaled dot-product attention:

$$\alpha_{ij} = \frac{\exp(\mathbf{q}_i^\top \mathbf{k}_j / \sqrt{d})}{\sum_j \exp(\mathbf{q}_i^\top \mathbf{k}_j / \sqrt{d})}, \tag{6}$$

where $d$ is the feature dimension. The fused representation for viewpoint $i$ is then computed as:

$$\mathbf{f}_{\text{fused},i} = \sum_j \alpha_{ij} \mathbf{v}_j. \tag{7}$$

In practice, this attention mechanism not only aggregates features but also dynamically reweights the contribution of each view, ensuring that unreliable or noisy views have a diminished impact on the final fused representation.

#### 3.3.2 Regularization for Attention Stability

To prevent the attention mechanism from overfitting to a single dominant viewpoint and ensure smooth multi-view integration, we introduce additional regularization techniques.

**Entropy Regularization:** To encourage the model to distribute attention weights more evenly and prevent the collapse of attention into a single dominant viewpoint, we introduce an entropy-based regularization loss:

$$\mathcal{L}_{\text{entropy}} = \sum_i \sum_j \alpha_{ij} \log \alpha_{ij}. \tag{8}$$

Minimizing this loss ensures that each viewpoint contributes evenly to the final fused representation.

**View Consistency Loss:** To enforce geometric consistency across viewpoints, we introduce a consistency loss that minimizes the variance of the fused features across viewpoints:

$$\mathcal{L}_{\text{consistency}} = \sum_i \left\| \mathbf{f}_{\text{fused},i} - \frac{1}{n} \sum_j \mathbf{f}_{\text{fused},j} \right\|_2^2. \tag{9}$$

This loss encourages all viewpoints to align towards a globally coherent representation.

In practice, the relative weights of $\mathcal{L}_{\text{entropy}}$ and $\mathcal{L}_{\text{consistency}}$ can be tuned to balance the trade-off between maintaining local view fidelity and enforcing global coherence.

#### 3.3.3 Multi-Head Self-Attention for Enhanced Feature Learning

To improve the robustness of the attention mechanism, we extend our self-attention fusion module using *multi-head self-attention* (MHSA) [26]. Instead of using a single set of query, key, and value projections, we use $H$ attention heads:

$$\mathbf{q}_i^h = W_q^h \mathbf{f}_i, \quad \mathbf{k}_i^h = W_k^h \mathbf{f}_i, \quad \mathbf{v}_i^h = W_v^h \mathbf{f}_i, \quad h \in \{1, ..., H\}. \tag{10}$$

Each head computes its own attention weights $\alpha_{ij}^h$ and corresponding fused feature:

$$\alpha_{ij}^h = \frac{\exp(\mathbf{q}_i^{h\top}\mathbf{k}_j^h/\sqrt{d})}{\sum_j \exp(\mathbf{q}_i^{h\top}\mathbf{k}_j^h/\sqrt{d})}. \tag{11}$$

The final fused feature is obtained by concatenating all head outputs and applying a linear projection:

$$\mathbf{f}_{\text{fused},i} = W_o\left[\mathbf{f}_{\text{fused},i}^1; \mathbf{f}_{\text{fused},i}^2; \ldots; \mathbf{f}_{\text{fused},i}^H\right]. \tag{12}$$

Multi-head self-attention allows the model to capture complementary information from different viewpoints, improving the robustness of the fused representation. This mechanism enables the network to attend to diverse aspects of the view-adaptive features simultaneously, thereby enhancing its ability to reconcile inconsistencies across views.

### 3.3.4 Final Loss Function

We adopt a typical photometric rendering loss to minimize the discrepancy between synthesized images and the ground-truth multi-view images:

$$\mathcal{L}_{\text{render}} = \sum_{\mathbf{r}\in\mathcal{R}} \left\|\widehat{C}(\mathbf{r}) - C(\mathbf{r})\right\|_2^2, \tag{13}$$

where $C(\mathbf{r})$ and $\widehat{C}(\mathbf{r})$ are the ground-truth and predicted pixel colors along ray $\mathbf{r}$, respectively.

The final loss function for training the self-attention fusion module combines the rendering loss $\mathcal{L}_{\text{render}}$, entropy regularization loss $\mathcal{L}_{\text{entropy}}$, and consistency loss $\mathcal{L}_{\text{consistency}}$:

$$\mathcal{L} = \lambda_{\text{render}}\mathcal{L}_{\text{render}} + \lambda_{\text{entropy}}\mathcal{L}_{\text{entropy}} + \lambda_{\text{consistency}}\mathcal{L}_{\text{consistency}}. \tag{14}$$

This formulation ensures that the fused representations remain geometrically coherent, stable, and adaptable to inconsistencies in multi-view inputs.

Note that the weights $\lambda_{\text{render}}$, $\lambda_{\text{entropy}}$, and $\lambda_{\text{consistency}}$ are empirically tuned to balance the contributions of each loss term, ensuring robust and stable training. Specifically, we empirically set the weight coefficients as $\lambda_{\text{render}} = 1.0$, $\lambda_{\text{entropy}} = 0.01$, and $\lambda_{\text{consistency}} = 0.1$. These values are chosen based on the following theoretical and empirical considerations: (1) *Rendering Loss Weight*: The rendering loss is the primary objective of the network, ensuring that synthesized images match the ground-truth multi-view images. Since this term directly supervises the learning of radiance fields, we set $\lambda_{\text{render}} = 1.0$ to maintain its dominant influence, as done in prior works on NeRF-based optimization [4]; (2) *Entropy Regularization Weight*: The entropy loss prevents overfitting to a single dominant viewpoint by encouraging uniform attention weight distribution. However, too strong a regularization may disrupt the selective aggregation of useful viewpoints. A small weight $\lambda_{\text{entropy}} = 0.01$ ensures that attention distribution remains balanced while still prioritizing reliable views, following the principles of entropy-maximizing regularization in self-attention [26]; and (3) *View Consistency Loss Weight*: This term enforces structural coherence across viewpoints. The weight is set relative to the rendering loss to allow slight adjustments for view consistency while not overpowering the primary photometric loss. The choice of $\lambda_{\text{consistency}} = 0.1$ aligns with prior work on enforcing geometric coherence in multi-view learning [30]. These values were determined through empirical tuning, ensuring stability and convergence without excessive regularization. The theoretical analysis of view-adaptive neural rendering is discussed in Sec. 3.5.

### 3.4 Training Strategy

Training is conducted in three phases to balance global and local learning:

1. **Global Pre-training:** We first train a single MLP (*without* view-adaptive branches) on all viewpoints for $T_{\text{pre}}$ iterations. This step initializes the parameters of the shared layers, ensuring they capture a global structure of the scene. This robust initialization is critical for the subsequent fine-tuning stages.
2. **view-adaptive Fine-Tuning:** Next, we expand the network to include $n$ view-adaptive MLP heads. We fine-tune these new branches independently, allowing each viewpoint to learn corrections specific to that view's inconsistencies. The shared layers remain partially frozen or are updated at a lower learning rate to preserve global knowledge. Empirically, updating the shared layers at a reduced learning rate (typically 1/10th of that used for the view-adaptive branches) helps maintain consistency across views.
3. **Self-Attention Fusion:** Finally, we enable the training of the self-attention fusion module, allowing it to combine view-adaptive features into a coherent representation. This stage refines the balance between viewpoint specialization and global geometric consistency. It is crucial to introduce the self-attention fusion module only after view-adaptive fine-tuning to avoid destabilizing the learned corrections.

### 3.5 Theoretical Analysis of view-adaptive Neural Rendering

In this section, we provide a theoretical justification for why our view-adaptive neural renderer can effectively handle view inconsistencies in multi-view images. The following analysis provides theoretical intuition on view-dependent inconsistencies. The experimental results in Secs. 4.3 and 4.6 further support this intuition. We formalize the problem and present a lemma demonstrating the impact of view inconsistency on NeRF-based reconstruction. We then discuss how our approach mitigates this issue, leading to improved 3D reconstruction.

**Definition 1 (View Inconsistency) [31].** Let $\mathcal{I} = \{I_1, I_2, \ldots, I_n\}$ be a set of multi-view images of a 3D object, where each $I_i$ corresponds to a viewpoint $\mathbf{v}_i$. The images are considered *view-consistent* if there exists a mapping function $\mathcal{T}$ that projects all images to a common 3D representation $\mathcal{V}$ such that:

$$\mathcal{T}(I_i, \mathbf{v}_i) = \mathcal{V}, \quad \forall i. \tag{15}$$

However, due to the inherent ambiguity in monocular view synthesis [32], images generated from a single image often suffer from *view inconsistency*, resulting in perturbations $\delta_i$:

$$\mathcal{T}(I_i, \mathbf{v}_i) = \mathcal{V} + \delta_i, \quad \text{where } \delta_i \neq 0. \tag{16}$$

Thus, standard NeRF-based methods trained on $\mathcal{I}$ learn a function that may inherit these inconsistencies, thereby degrading the quality of the 3D reconstruction.

**Lemma 1 (Effect of View Inconsistency under Averaged Aggregation).** Assume that the per-view projection satisfies $\mathcal{T}(I_i, \mathbf{v}_i) = \mathcal{V} + \delta_i$ for $i = 1, \ldots, n$. Define the aggregated estimate

$$\hat{\mathcal{V}} := \frac{1}{n} \sum_{i=1}^{n} \mathcal{T}(I_i, \mathbf{v}_i). \tag{17}$$

Then the reconstruction error satisfies

$$\|\hat{\mathcal{V}} - \mathcal{V}\|_2 \leq \frac{1}{n} \sum_{i=1}^{n} \|\delta_i\|_2. \tag{18}$$

This lemma analyzes an idealized aggregation that captures how per-view perturbations can accumulate under multi-view supervision.

Our view-adaptive neural renderer addresses this issue by incorporating dedicated corrective modules for each viewpoint. These view-adaptive branches learn to mitigate the individual perturbations $\delta_i$, thereby reducing the overall reconstruction error and improving the fidelity of the generated 3D models.

**Proof.** For analysis, we consider a simplified estimator defined as the average of the per-view projected representations:

$$\hat{\mathcal{V}} \ := \ \frac{1}{n} \sum_{i=1}^{n} \mathcal{T}(I_i, \mathbf{v}_i). \tag{19}$$

Since $\mathcal{T}(I_i, \mathbf{v}_i) = \mathcal{V} + \delta_i$, we obtain:

$$\hat{\mathcal{V}} = \mathcal{V} + \frac{1}{n} \sum_{i=1}^{n} \delta_i. \tag{20}$$

By the triangle inequality,

$$\|\hat{\mathcal{V}} - \mathcal{V}\|_2 = \left\| \frac{1}{n} \sum_{i=1}^{n} \delta_i \right\|_2 \leq \frac{1}{n} \sum_{i=1}^{n} \|\delta_i\|_2. \tag{21}$$

Thus, as view inconsistencies $\|\delta_i\|_2$ increase, the reconstruction error increases, and is bounded above by the average magnitude of the perturbations. □

**Theorem 1 (view-adaptive Learning Reduces View Inconsistency).** Assume that view-adaptive learning reduces each per-view perturbation $\delta_i$ to a residual $\delta_i'$ such that $\|\delta_i'\|_2 \leq (1-\lambda_{\text{adapt}})\|\delta_i\|_2$ for some $0 < \lambda_{\text{adapt}} < 1$. Then the reconstruction error satisfies:

$$\|\hat{\mathcal{V}}_{\text{ours}} - \mathcal{V}\|_2 \leq (1 - \lambda_{\text{adapt}}) \cdot \frac{1}{n} \sum_{i=1}^{n} \|\delta_i\|_2, \tag{22}$$

where $\lambda_{\text{adapt}} > 0$ is a corrective factor learned through view-adaptive adaptation. This inequality indicates that our approach reduces the effective reconstruction error by a multiplicative factor that is proportional to the average view inconsistency.

**Corollary 1 (Effect of Self-Attention Fusion) [26].** Assume the fusion weights are produced by a softmax normalization, so that $\beta_i \geq 0$ and $\sum_{i=1}^{n} \beta_i = 1$, and the fused estimate is $\hat{\mathcal{V}}_{\text{fused}} = \sum_{i=1}^{n} \beta_i \hat{\mathcal{V}}_{\text{ours},i}$. Then the reconstruction error is bounded by:

$$\|\hat{\mathcal{V}}_{\text{fused}} - \mathcal{V}\|_2 \leq \sum_{j=1}^{n} \beta_j \|\hat{\mathcal{V}}_{\text{ours},j} - \mathcal{V}\|_2 \leq \max_j \|\hat{\mathcal{V}}_{\text{ours},j} - \mathcal{V}\|_2. \tag{23}$$

This shows that attention-based fusion can suppress outlier views by aggregating information through a weighted averaging mechanism.

# 4 Experiments

This section experimentally validates the effectiveness of our view-adaptive neural renderer for single-image 3D reconstruction. We begin by describing the datasets and evaluation metrics (Sec. 4.1), followed by the implementation details of our framework (Sec. 4.2). We then compare our method with baseline NeRF-based models to evaluate the impact of view-adaptive training (Sec. 4.3) and analyze how different loss functions impact both reconstruction quality and computational efficiency (Sec. 4.4). Subsequently, we compare our proposed framework with recent state-of-the-art 2D-to-3D reconstruction methods (Sec. 4.5). Finally, we investigate the robustness of our method regarding the number of available viewpoints (Sec. 4.6) and provide an ablation study to analyze the contribution of each module (Sec. 4.6).

## 4.1 Performance Metrics and Datasets

To compare the accuracy of novel view synthesis, we calculated three metrics: *Peak Signal-to-Noise Ratio* (PSNR) [33], *Structural Similarity Index Measure* (SSIM) [33], and *Learned Perceptual Image Patch Similarity* (LPIPS) [34]. We also evaluated the accuracy of 3D reconstruction using two metrics, following recent related work: *Chamfer Distances* (CD) [35] and *Volume Intersection over Union* (IoU) [36].

The multi-view images used for 3D reconstruction were generated using the recent SyncDreamer method [2]. While these images are synthesized to simulate real-world camera viewpoints, some artifacts or noise may persist due to the limitations inherent in monocular image generation. Such artifacts are taken into account when interpreting the reconstruction results. Following recent larger-scale single-image-to-3D methods [37–39], we utilized 50 3D object instances from the Google Scanned Object (GSO) [40] for evaluation. These selected instances were not used in the training process of the multi-view generation method. This ensures that the evaluation is conducted on unseen data, providing an unbiased assessment of our method.

Unless otherwise specified, multi-view images were rendered for each object at 256×256 for four view-points (0°, 90°, 180°, and 270°), and these multi-view images are used to generate 3D reconstruction. For the analysis in Sec. 4.6, we additionally render 8, 12, and 16 views uniformly sampled in azimuth. *InstantNGP* [6] and *NeuS* [5] were employed as baseline neural renderers for 3D reconstruction.

## 4.2 Implementation Details

We validate our method using two representative *NeRF*-based models, **NeuS** [5] and **InstantNGP** [6], highlighting the broad applicability of our framework.

- **Network Configuration:** The shared MLP has 4 layers with 256 hidden units, while each view-adaptive MLP has 3 layers with 128 units. The self-attention block uses a multi-head attention mechanism (we set the number of heads to 4 by default). These hyperparameters were empirically chosen to balance model capacity with computational efficiency.
- **Positional Encoding:** We use a standard positional encoding [4] with 6 frequency bands for $\mathbf{x}$ and 4 for $\mathbf{\Theta}$. These settings can be adjusted based on scene complexity.
- **Optimization:** We use Adam with a learning rate of $5 \times 10^{-4}$ for the shared MLP during pre-training. During the view-adaptive fine-tuning, we reduce the learning rate of the shared layers by a factor of 10 and keep the view-adaptive layers at $5\times10^{-4}$. The total number of iterations is split into ($100k$, $200k$) for the pre-training and fine-tuning stages, respectively. The 200k fine-tuning stage includes both view-adaptive branch tuning and subsequent enabling of the self-attention fusion module. These optimization settings were tuned to ensure stable convergence while effectively learning both global and view-adaptive features.
- **Data Preprocessing and Augmentation:** Input images are normalized to the range $[0, 1]$and resized to 256×256 pixels. No additional data augmentation was applied, as the synthetic multi-view images are generated with consistent pre-processing.
- **Random Seed and Reproducibility:** To ensure reproducibility, we fix the random seed to 42 for all experiments.
- **Hardware and Batch Size:** All experiments run on a single NVIDIA RTX 3090 GPU with a batch size of $1,024$ rays per iteration. Our implementation is based on PyTorch 1.9 with CUDA 11.1, and the framework scales efficiently to multi-GPU setups if needed.

## 4.3 Comparisons with Baselines

To validate the proposed view-adaptive neural renderer, we compared performance between neural renderers with and without view-adaptive training.

**Table 1** Quantitative comparison of 3D reconstruction performance using *NeRF*-based baseline methods with and without view-adaptive training.

| **Method/Metric** | **PSNR** (↑) | **SSIM** (↑) | **LPIPS** (↓) | **CD** (↓) | **IoU** (↑) |
|---|---|---|---|---|---|
| Original *InstantNGP* [6] | 20.04 | 0.768 | 0.159 | 0.0230 | 0.5169 |
| View-adaptive *InstantNGP* | 21.89 | 0.813 | 0.145 | 0.0142 | 0.6597 |
| Original *NeuS* [5] | 19.76 | 0.790 | 0.153 | 0.0168 | 0.6268 |
| View-adaptive *NeuS* | **22.74** | **0.833** | **0.130** | **0.0122** | **0.7015** |

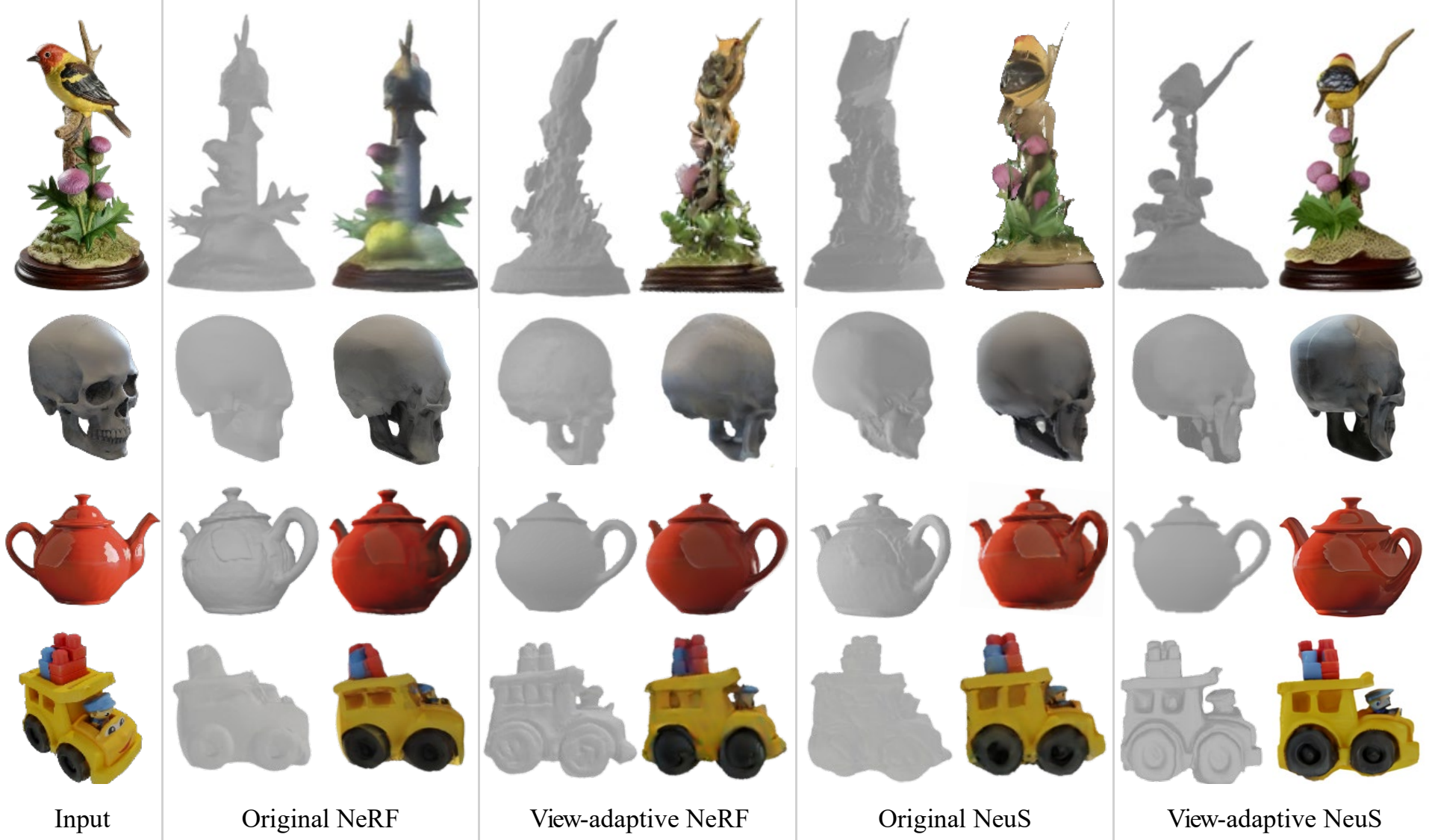

**Fig. 3** Qualitative comparison of 3D reconstructions produced by the proposed view-adaptive neural renderer and traditional *NeRF*-based methods, highlighting improvements in detail and consistency.

The results are summarized in Table 1. Comparing *InstantNGP* [6] and *NeuS* [5] with and without the view-adaptive neural renderer shows significant accuracy improvements across all metrics with view-adaptive training. In particular, improvements in PSNR and SSIM indicate sharper and more structurally coherent reconstructions, while the reductions in LPIPS and Chamfer Distance, along with increased IoU, reflect enhanced perceptual quality and geometric accuracy. These results demonstrate that the proposed view-adaptive neural renderer enables more accurate 3D reconstruction and novel-view synthesis from view-inconsistent multi-view images.

Figure 3 shows 3D meshes and images synthesized from novel viewpoints, compared to those generated by baseline methods. It is shown that 3D reconstructions of neural renderers without view-adaptive training are over-smoothed and lack details due to the multi-view inconsistency. However, when the network is trained view-adaptively, 3D reconstruction with *NeuS* and *InstantNGP* are more visually continuous and represent fine details. These results confirm our approach compensates for input view inconsistencies, enhancing reconstruction quality.

The quantitative and qualitative improvements observed in our experiments underscore the robustness of our framework under challenging conditions. The ability to correct viewpoint-adaptive discrepancies not only enhances local detail recovery but also preserves the global structural integrity of the reconstructed 3D models. This is particularly significant when dealing with artifacts or noise inherent in multi-view images generated from monocular inputs. The integration of view-adaptive neural renderers ensures that individual viewpoints are refined independently, reducing error propagation and enhancing overall consistency. Furthermore, the self-attention fusion mechanism adaptively aggregates information across views, reinforcing global coherence. By dynamically reweighting the contribution of each view, our method effectively leverages complementary features and suppresses unreliable information. This design choice proves to be especially advantageous when the number of input views is limited, as it compensates for the reduced overlap in geometric information.

Overall, our results highlight that the proposed technique not only outperforms traditional NeRF-based approaches but also offers a practical trade-off between accuracy and computational efficiency. The demonstrated improvements in both reconstruction fidelity and rendering quality suggest that our method is well-suited for real-world applications, such as virtual reality, augmented reality, and digital content creation, where robust and efficient 3D generation from single images is critical.

## 4.4 Evaluation of View-adaptive Rendering

Our method achieves accurate 3D reconstruction and novel view synthesis using only the rendering loss, thanks to the view-adaptive training. Here, we evaluated the performance of the proposed view-adaptive neural renderer under different loss functions. Specifically, we compared the accuracy of 3D reconstruction when using the rendering loss alone, the SDS loss alone, and a combination of both. In our experiments, *NeuS* was used as the baseline for 3D reconstruction, and the results are summarized in Table 2.

Rendering loss provides accurate 3D reconstruction in faster processing time, assuming that the multi-view images are visually consistent. In contrast, traditional *NeRF*-based methods that rely solely on rendering loss often suffer from reduced accuracy due to visual discontinuities in multi-view images. However, the view-adaptive neural renderer demonstrates that the rendering loss is sufficient to compensate for these discontinuities, achieving robust and visually coherent reconstructions.

Using both the rendering loss and the SDS loss yields the highest reconstruction accuracy, yet the processing time is significantly increased compared to using the rendering loss alone. Notably, the results achieved using only the rendering loss are nearly comparable to those using both losses, with only a slight reduction in accuracy but an approximately 87% decrease in training time. Furthermore, training a view-adaptive neural renderer using only the SDS loss did not improve performance in our setting, suggesting that SDS-only optimization may conflict with view-adaptive specialization without photometric supervision.

These findings highlight several key strengths of our approach. First, our technique effectively isolates and compensates for view-adaptive inconsistencies, which enables accurate 3D reconstructions even under less-than-ideal multi-view inputs. Second, the ability to achieve nearly state-of-the-art performance using only rendering loss underscores the efficiency of our view-adaptive training strategy, making the method highly practical for applications where computational resources or training time are limited. Lastly, the minimal accuracy trade-off when omitting the more complex SDS loss demonstrates that our approach inherently stabilizes the reconstruction process by leveraging the inherent advantages of view-adaptive adaptation. Overall, these results not only validate the robustness of our method but also emphasize its practical significance in real-world scenarios where multi-view consistency cannot be guaranteed.

**Table 2** Quantitative comparison of view-adaptive training using different loss functions. **Bold** indicates the best performance, and underline denotes the second-best result for each metric.

| Loss Function | Chamfer Distance (↓) | | Volume IoU (↑) | | Time (min) | |
|---|---|---|---|---|---|---|
| | Original | View-adaptive | Original | View-adaptive | Original | View-adaptive |
| Rendering Loss | 0.0168 | 0.0122 | 0.6268 | 0.7015 | **5** | 7 |
| SDS Loss | 0.0154 | 0.0156 | 0.6570 | 0.5925 | 40 | 45 |
| Rendering + SDS Loss | 0.0150 | **0.0109** | 0.6728 | **0.7042** | 48 | 54 |

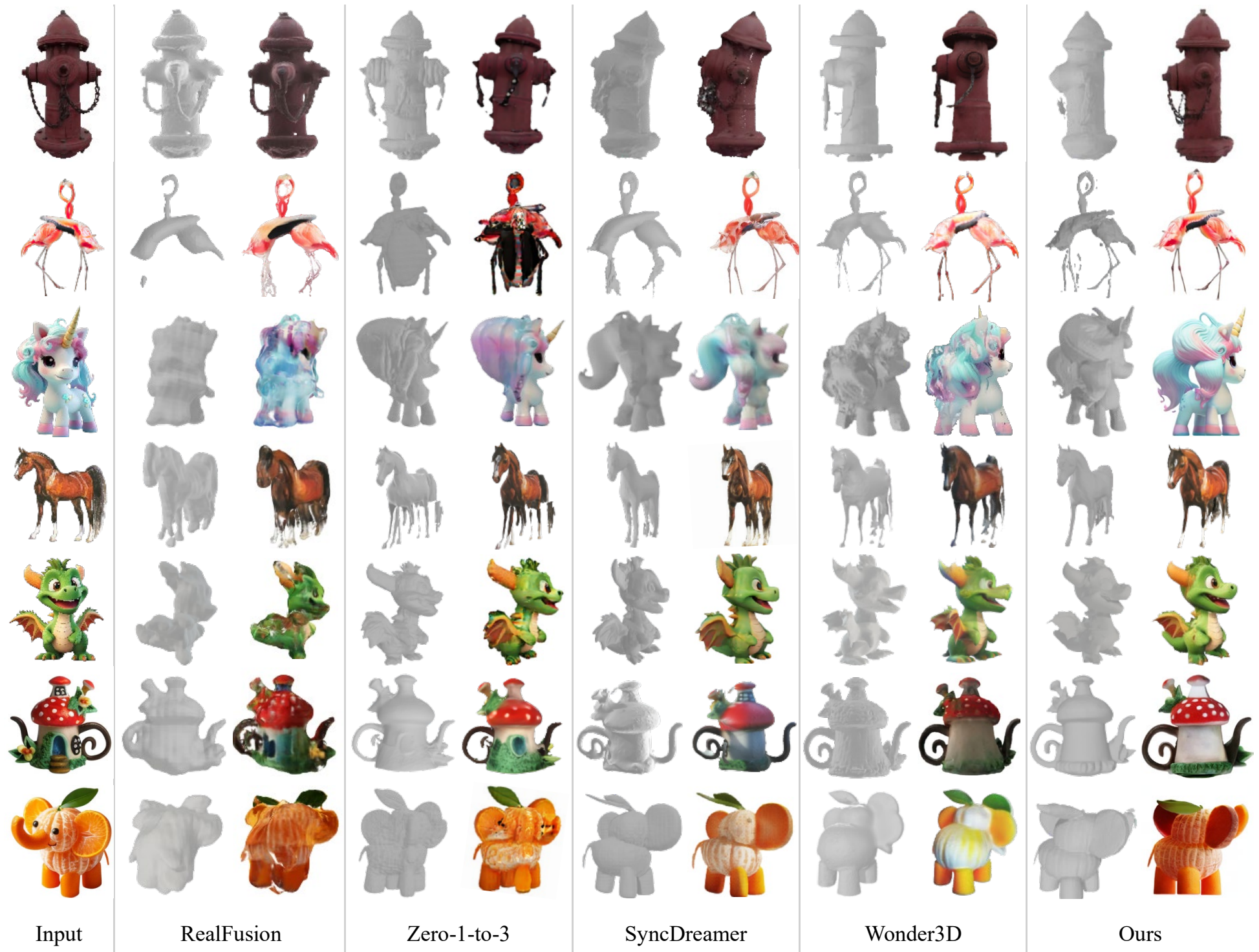

**Fig. 4** Qualitative comparisons with state-of-the-art single-image-to-3D methods under a shared *NeuS* backend: RealFusion [3], Zero-1-to-3 [19], Wonder3D [20], and SyncDreamer [2]. The figure illustrates that our method yields more accurate and detailed 3D reconstructions from inconsistent multi-view images.

## 4.5 Comparisons with the State-of-the-arts Methods

To validate the effect of the proposed view-adaptive training under a share backend setting, we compare the performance with recent single image to 3D generation methods: *RealFusion* [3], *Zero-1-to-3* [19], *SyncDreamer* [2], and *Wonder3D* [20]. *Zero-1-to-3* [19] utilizes a diffusion model to learn how to control the camera extrinsic. *Zero-1-to-3* takes a single image and the relative pose of the camera as input to generate a novel view image corresponding to the given pose. Then, the generated multi-view images are utilized for 3D reconstruction using the Score Jacobian Chaining (SJC) loss function. *RealFusion* [3] is a single image to 3D generation method that leverages the 2D diffusion model to learn 3D geometry and appearance by sampling images from different viewpoints. *Wonder3D* [20] is a method for generating high-resolution texture meshes from a single image, using a cross-domain diffusion model to generate multi-view normal maps and color images to improve the quality of 3D reconstructions. Here,

**Table 3** Comparison with state-of-the-art 2D-to-3D methods under a shared *NeuS* backend on 300 GSO objects. The proposed view-adaptive training improves reconstruction quality over the original *NeuS* using the same SyncDreamer-generated inputs [2].

| Method/Metric | PSNR (↑) | SSIM (↑) | LPIPS (↓) | CD (↓) | IoU (↑) |
|---|---|---|---|---|---|
| RealFusion [3] | 14.98 | 0.714 | 0.292 | 0.0758 | 0.3562 |
| Zero-1-to-3 [19] | 18.62 | 0.770 | 0.174 | 0.0255 | 0.5884 |
| SyncDreamer [2] | 19.76 | 0.790 | 0.153 | 0.0168 | 0.6268 |
| Wonder3D [20] | 20.51 | 0.803 | 0.148 | 0.0159 | 0.6512 |
| Ours | **22.74** | **0.833** | **0.130** | **0.0122** | **0.7015** |

we used the same evaluation metrics and dataset as described in Sec. 4.1. For multi-view image generation of our method, we used *SyncDreamer* [2]. We used *NeuS* as a shared reconstruction backend because it is a strong surface-oriented reconstruction method and achieves stronger reconstruction performance in Sec. 4.3. This provides a consistent reconstruction condition for the shared-backend evaluation

Table 3 shows a quantitative comparison with state-of-the-art 2D-to-3D methods under a shared reconstruction backend. For clarity, all numbers in Table 3 are obtained using the same 3D reconstruction backend (*i.e.*, *NeuS*) to provide a consistent comparison across different multi-view generators. Therefore, the 'SyncDreamer' row in Table 3 corresponds to reconstructing with *NeuS* from SyncDreamer-generated multi-view images, which is identical to the 'Original NeuS' setting in Table 1. Likewise, the 'Ours' row in Table 3 corresponds to applying our view-adaptive training on top of *NeuS* under the same SyncDreamer inputs, matching the 'View-adaptive NeuS' setting in Table 1. Hence, the improvements from 'SyncDreamer' to 'Ours' in Table 3 should be interpreted as the gain from view-adaptive training under the same generated inputs and reconstruction backend. The results show that the proposed view-adaptive training improves reconstruction quality over the original *NeuS* when using the same multi-view inputs. This indicates that the proposed view-adaptive renderer can better compensate for partial view inconsistencies in generated multi-view images under the shared-backend setting.

Figure 4 shows 3D reconstruction comparisons with the recent methods. *RealFusion* [3] exhibits suboptimal multi-view consistency and fails to produce visually detailed images for complex objects. *Zero-1-to-3* [19] generates more visually detailed images than *RealFusion*, but results in overly smoothed 3D reconstructions. These results indicate that indirect loss functions, such as SDS and Score Jacobian Chaining (SJC), are limited in generating visually continuous 3D reconstructions from inconsistent multi-view images. Additionally, *Wonder3D* [20] assumes a fixed focal length for training (*e.g.*, 35mm), leading to distortions in images with varying focal lengths. In contrast, despite using the same multi-view images as *SyncDreamer* [2], our neural renderer applies view-adaptive learning, effectively addressing discontinuities and generating the most visually accurate 3D reconstruction. In addition, the proposed method can generate semantically consistent images that are coherent with the input image, as well as multi-view images that maintain consistency in color and geometry.

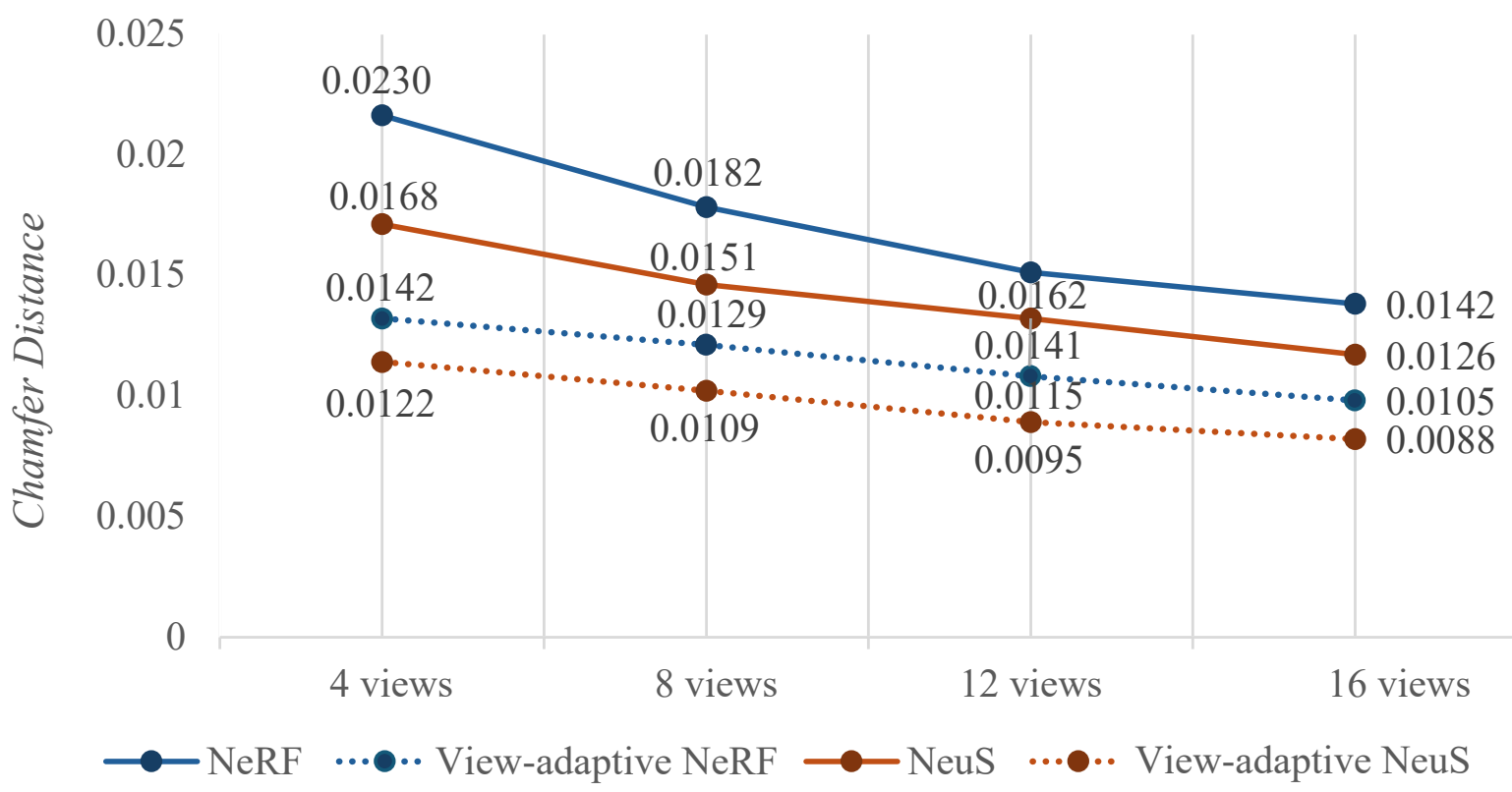

**Fig. 5** Effect of the number of views on 3D reconstruction quality. The figure compares the performance of the proposed view-adaptive neural renderer with the standard *NeRF*-based methods across varying view counts.

These consistent improvements under the shared-backend setting have several practical implications. First, by explicitly modeling view-dependent variations, our method overcomes a key limitation of conventional *NeRF*-based models that assume global consistency across all viewpoints. This yields improved reconstruction fidelity even with localized inconsistencies in input images. Second, achieving high-quality results with a simple rendering loss, rather than more computationally intensive indirect loss functions, enhances practicality for real-world applications constrained by training time and resources. Finally, the enhanced detail preservation and geometric accuracy of our method highlight its potential for applications in virtual reality, augmented reality, and high-fidelity digital content creation. Overall, these comparisons emphasize the quantitative advantages and practical strengths of our view-adaptive neural renderer in efficiency and robustness with imperfect multi-view inputs.

## 4.6 Performance by Number of Viewpoints

To compare the performance depending on the number of multi-view images used for 3D reconstruction, we evaluated the performance of neural renderers with and without view-adaptive training for different quantities of views. We used CD as a metric to evaluate the accuracy of the 3D reconstruction, varying the number of viewpoints to 4, 8, 12, and 16 views.

The results are summarized in Fig. 5. It is shown that when the number of views is 4, the performance of view-adaptive neural renderers has the highest performance gain over the corresponding baseline methods, and the performance gain decreases as the number of views increases. This indicates that view-adaptive training is particularly beneficial in scenarios with limited input views. Without view-adaptive training, performance improves distinctly as the number of views increases. This is because, given enough multi-view images, the error caused by partial view inconsistency is compensated for by information from different views. In contrast, when a view-adaptive neural renderer is employed, the performance of the 3D reconstruction is not significantly affected by the number of generated multi-view images. In fact, even with only a few views, the reconstruction quality remains robust. The results show that the proposed view-adaptive neural renderer can effectively handle the view inconsistency of multi-view images to generate accurate 3D reconstruction, even using a few images. It is demonstrated that view-adaptive training enables the generation of visually consistent 3D reconstructions despite inherent view discontinuities.

**Table 4** Component ablation study under the same *NeuS* setting using multi-view images generated by SyncDreamer [2]. The baseline corresponds to reconstruction with the original *NeuS*, and each row cumulatively adds the indicated component to the previous setting.

| **Method/Metric** | **PSNR** (↑) | **SSIM** (↑) | **LPIPS** (↓) | **CD** (↓) | **IoU** (↑) |
|---|---|---|---|---|---|
| Baseline | 19.76 | 0.790 | 0.153 | 0.0168 | 0.6268 |
| + View-adaptive branches | 21.94 | 0.820 | 0.139 | 0.0136 | 0.6788 |
| + Self-attention fusion | 22.34 | 0.827 | 0.135 | 0.0129 | 0.6902 |
| + Entropy regularization | 22.55 | 0.830 | 0.132 | 0.0126 | 0.6955 |
| + Consistency loss (Ours) | **22.74** | **0.833** | **0.130** | **0.0122** | **0.7015** |

These findings are particularly significant as they highlight the strength of our approach in resource-constrained settings, where acquiring numerous multi-view images may be impractical. The robustness of the reconstruction quality with a limited number of views not only reduces data collection costs but also accelerates the overall processing time. Moreover, by mitigating the dependency on a high number of input views, our method demonstrates enhanced adaptability and generalization across diverse scenarios. This robustness underscores the practical value of our view-adaptive neural renderer in real-world applications where multi-view consistency is challenging to achieve.

## 4.7 Ablation Study.

To analyze the contribution of each component, we conduct an ablation study under the same *NeuS* setting using multi-view images generated by SyncDreamer [2]. We use the same evaluation metrics and dataset as described in Sec. 4.1. As shown in Table 4, the baseline corresponds to the original NeuS reconstruction, and each row cumulatively adds the indicated component to the previous setting. Adding view-adaptive branches leads to the largest improvement across all metrics. In particular, the increase in PSNR and SSIM and the decrease in LPIPS indicate that viewpoint-specific branches effectively correct local texture variations and appearance discontinuities, thereby improving novel-view rendering quality. At the same time, the decrease in CD and the increase in IoU show that these viewpoint-specific corrections are not limited to appearance refinement, but also directly contribute to 3D reconstruction fidelity by reducing over-smoothed geometry and local shape distortions.

Adding self-attention fusion further improves the performance. This is because, rather than simply combining independently learned view-adaptive features from each viewpoint, the self-attention module can emphasize relatively reliable views while reducing the influence of noisy or inconsistent views. As a result, LPIPS and CD are further reduced, and IoU is improved, indicating that both perceptual consistency and geometric coherence are enhanced. Entropy regularization prevents the attention distribution from being overly concentrated on a specific view, allowing multi-view information to be used more stably. This leads to stable improvements in view-level consistency metrics such as SSIM and LPIPS. Finally, the consistency loss prevents viewpoint-specific adaptation from becoming overly independent and encourages the model to preserve a shared 3D structure. This results in additional improvements in CD and IoU, and the final full model achieves the best performance in both rendering quality and geometry accuracy. These progressive improvements show that the view-adaptive branches, self-attention fusion, entropy regularization, and consistency loss complement different aspects of reconstruction quality and jointly contribute to the final performance.

## 4.8 Discussion and Analysis

In this section, we provide a more in-depth analysis of our experimental results by linking the observed performance gains to the technical background of neural rendering and view inconsistency.

**1) Impact of view-adaptive Adaptation.** As shown in Table 1 and Figure 3, the proposed view-adaptive neural renderer consistently outperforms traditional NeRF-based approaches in both PSNR and SSIM. We attribute these improvements to the explicit modeling of viewpoint-dependent variations. In standard NeRF pipelines, a single MLP attempts to encode all viewpoints, assuming a globally consistent scene representation. However, when the input multi-view images contain partial inconsistencies (e.g., small misalignments or texture variations), a single global network struggles to reconcile these discrepancies. In contrast, our method allows each viewpoint to learn specialized corrections (via dedicated MLP heads), effectively mitigating local distortions without corrupting the shared representation. This targeted adaptation prevents error propagation and ensures that local inconsistencies are isolated and corrected at the source.

**2) Role of Self-Attention Fusion.** The self-attention fusion module (Section 3.3) further refines the aggregated features by adaptively weighting reliable viewpoints and suppressing noisy or inconsistent ones. From Table 2, one can see that combining our view-adaptive approach with self-attention leads to more robust and coherent 3D reconstructions in terms of Chamfer Distance (CD) and IoU. This agrees with our theoretical analysis (Theorem 1), which suggests that learned corrective factors reduce the effective influence of inconsistent views. Moreover, the dynamic weighting provided by the self-attention mechanism helps balance the contributions from all views, further enhancing the overall reconstruction quality.

**3) Rendering Loss vs. SDS Loss.** Table 2 shows an interesting trade-off: using only rendering loss achieves nearly the same accuracy as combining rendering and SDS losses, but at a significantly lower computational cost. In typical settings, SDS loss helps to impose additional priors on geometry or texture, particularly when multi-view images are very noisy or insufficient. However, because our view-adaptive renderer already addresses local distortions by design, the marginal benefit from SDS is reduced. Hence, practitioners can choose to disable SDS when faster training is preferred and multi-view images are moderately reliable. This result underscores the efficiency of our method, achieving near-optimal performance while substantially reducing training time.

**Table 5** Generalizability analysis of the proposed view-adaptive training using multi-view inputs generated by various 2D-to-3D methods under the same NeuS reconstruction backend.

| Method/Metric | PSNR (↑) | SSIM (↑) | LPIPS (↓) | CD (↓) | IoU (↑) |
|---|---|---|---|---|---|
| Zero-1-to-3 [19] | 18.62 | 0.770 | 0.174 | 0.0255 | 0.5884 |
| Zero-1-to-3 [19] + Ours | **20.68** | **0.804** | **0.156** | **0.0199** | **0.6467** |
| Wonder3D [20] | 20.51 | 0.803 | 0.148 | 0.0159 | 0.6512 |
| Wonder3D [20] + Ours | **21.86** | **0.824** | **0.136** | **0.0134** | **0.6869** |

**4) Comparisons with Prior Methods.** Figure 4 and Table 3 provide qualitative and quantitative comparisons with recent single-image-to-3D approaches, including RealFusion [3], Zero-1-to-3 [19], and Wonder3D [20]. Under the shared *NeuS* reconstruction backend, the proposed view-adaptive training improves geometry fidelity and view consistency when applied to the same SyncDreamer-generated inputs. Specifically, RealFusion sometimes struggles with high-frequency details due to indirect optimization constraints, while Zero-1-to-3 often produces over-smoothed geometries. In contrast, our method utilizes the synergy of shared/global and view-adaptive/local representations, yielding sharper reconstructions even when the multi-view images are imperfect. This synergy is pivotal in reconciling inconsistencies that are otherwise challenging for methods relying solely on indirect loss functions.

**5) Generalizability to Various Multi-view Generators.** Following the controlled evaluation protocol in Sec. 4.5, we further examine the generalizability of the proposed view-adaptive training by using multi-view images generated by Zero-1-to-3 [19] and Wonder3D [20] under the same *NeuS* reconstruction backend. As shown in Table 5, applying the proposed view-adaptive training consistently improves reconstruction quality over the corresponding original *NeuS* for both generators. This consistent trend suggests that our method is not limited to a specific multi-view generator, but can compensate for view-dependent inconsistencies in multi-view inputs generated by various 2D-to-3D methods. Overall, the proposed view-adaptive neural renderer can be flexibly applied to multi-view inputs from various generators while maintaining consistent reconstruction improvements.

**6) Sensitivity to Number of Views.** Figure 5 examines performance with varying numbers of input views. When the number of views is as low as four, baselines suffer from blurred reconstructions due to insufficient overlapping geometry. Our proposed approach still maintains superior accuracy in such scenarios, indicating that view-adaptive heads effectively compensate for limited or noisy viewpoints. As the number of views increases, the performance gap narrows, but our method remains consistently better or comparable due to adaptive fusion via multi-head self-attention. These observations suggest that our approach is particularly valuable in applications where only a few views are available.

**7) Generalizability to 3D Representations.** Although our experiments are conducted on *InstantNGP* [6] and *NeuS* [5], the core idea is not tied to a specific neural representation, but rather to mitigating viewpoint-dependent discrepancies in generated multi-view inputs during reconstruction. Therefore, view-adaptive adaptation can also be extended to reconstruction frameworks based on other 3D representations.

For example, in an explicit representation such as 3D Gaussian Splatting (3DGS) [41], an extension would be to keep the canonical Gaussian geometry, such as position, scale, and rotation, shared across viewpoints. Instead of learning view-specific geometry, view-conditioned adaptation can be applied to appearance-related components, such as color, spherical harmonics coefficients, learned appearance features, or a lightweight residual rendering module. This design can preserve the consistency of explicit Gaussian geometry while alleviating viewpoint-dependent appearance discrepancies in generated multi-view inputs. However, adding view-specific appearance attributes or residual modules may increase memory or parameter cost as the number of viewpoints grows. Therefore, a 3DGS-based extension can be designed by preserving shared explicit geometry and selectively applying view-conditioned adaptation to appearance-related components. This suggests the generalizability of the proposed view-adaptive framework to 3D reconstruction frameworks beyond NeRF-style renderers.

**Summary.** Overall, the experimental findings validate the core idea that *explicit view-adaptive adaptation* is crucial for handling partial inconsistencies in monocularly generated multi-view images. By employing a combination of specialized MLP layers and self-attention-based feature fusion, the proposed framework not only improves 3D reconstruction fidelity but also reduces the reliance on heavier indirect loss functions (e.g., SDS). This work demonstrates a practical balance between reconstruction accuracy and computational efficiency. Future work can further investigate adaptive view selection or improved camera parameter estimation to enhance robustness under large viewpoint gaps and complex backgrounds.

# 5 Conclusion

We presented a novel framework for single-image-to-3D generation that addresses the limitations of inconsistent multi-view images through a specialized view-adaptive neural renderer. Unlike traditional *NeRF*-based methods, our approach achieved high reconstruction accuracy, even with partial discontinuities in generated multi-view images. This was facilitated by a combination of shared and viewpoint-adaptive MLP layers with a self-attention fusion mechanism.

Experimental results demonstrate that the proposed method improves reconstruction quality over the corresponding original reconstruction baselines using the same generated multi-view inputs, while reducing the reliance on computationally intensive indirect supervision. Nevertheless, as our framework builds on *NeRF* variants, it retains some limitations, such as reduced reliability when the multi-view inputs are excessively noisy or accompanied by highly complex backgrounds. Future work will explore more robust approaches for handling large viewpoint gaps and cluttered scenes, as well as optimizing camera parameter estimation in tandem with the view-adaptive rendering process.

# Declarations

## Author contribution

U-Chae Jun: Software, Validation, Writing−original draft preparation, Jaeeun Ko: Software, Validation, Writing−original draft preparation, Jiwoo Kang: Conceptualization, Methodology, Investigation, Writing−review and editing, Supervision.

## 5.1 Funding

Not applicable.

## Data availability

We used publicly accessible datasets online: Objaverse in https://objaverse.allenai.org and Google Scanned Object (GSO) in https://goo.gle/scanned-objects.

## Conflict of interest

The authors declare no competing interests.

## Ethics approval

Not applicable.